\documentclass[runningheads]{llncs}
\usepackage[T1]{fontenc}
\usepackage{graphicx}
\usepackage{booktabs}%
\usepackage[misc]{ifsym}

\usepackage{hyperref}
\usepackage{footmisc}
\usepackage[dvipsnames,svgnames]{xcolor}
\usepackage{amsfonts}
\usepackage{amssymb}
\usepackage{amsmath}
\usepackage[nameinlink,capitalize]{cleveref}
\usepackage{caption}
\usepackage{subcaption}
\usepackage{tabularx}

\usepackage[final]{pdfpages}

\definecolor{navy}{rgb}{0.1, 0.1, 0.8}
\definecolor{gray}{rgb}{0.4, 0.4, 0.4}
\definecolor{olive}{rgb}{0.1, 0.5, 0.1}
\definecolor{ruby}{rgb}{0.8, 0.1, 0.3}
\definecolor{darkpastelgreen}{rgb}{0.01, 0.75, 0.24}
\definecolor{celestialblue}{rgb}{0.29, 0.59, 0.82}
\definecolor{coral}{rgb}{1.0, 0.5, 0.31}
\definecolor{blue}{rgb}{0.23, 0.44, 0.62}
\definecolor{Goldenrod}{rgb}{0.8,0.8,0}

\usepackage{soul}
\usepackage[colorinlistoftodos,prependcaption,textsize=tiny,textwidth=35mm]{todonotes}
\usepackage{xspace}

\newcommand{\eat}[1]{}

\newcommand{\pioc}[1]{#1}

\newcommand{\bmh}{BMH\xspace}
\newcommand{\bmhp}{BMH-P\xspace}
\newcommand{\bmhk}{BMH-K\xspace}

\renewcommand{\vec}[1]{\ensuremath{\overrightarrow{#1}}}

\begin{document}
\title{What Drives Online Popularity: Author, Content or Sharers? Estimating Spread Dynamics with Bayesian Mixture Hawkes}
\titlerunning{Bayesian Mixture Hawkes}
%
\author{Pio Calderon$^{(\textrm{\Letter})}$\orcidID{0000-0002-8747-8917} \and
Marian-Andrei Rizoiu\orcidID{0000-0003-0381-669X}}
\authorrunning{P. Calderon and M.A. Rizoiu}
%
\institute{University of Technology Sydney, Australia
\email{piogabrielle.b.calderon@student.uts.edu.au}\\
\email{marian-andrei.rizoiu@uts.edu.au}}

\tocauthor{Pio Calderon,Marian-Andrei Rizoiu}
\toctitle{What Drives Online Popularity: Author, Content or Sharers? Estimating Spread Dynamics with Bayesian Mixture Hawkes}

\maketitle              
\begin{abstract}
The spread of content on social media is shaped by intertwining factors on three levels: the source, the content itself, and the pathways of content spread. 
At the lowest level, the popularity of the sharing user determines its eventual reach.
However, higher-level factors such as the nature of the online item and the credibility of its source also play crucial roles in determining how widely and rapidly the online item spreads.
In this work, we propose the Bayesian Mixture Hawkes (\bmh) model to jointly learn the influence of source, content and spread. We formulate the \bmh model as a hierarchical mixture model of separable Hawkes processes, accommodating different classes of Hawkes dynamics and the influence of feature sets on these classes. We test the \bmh model on two learning tasks, cold-start popularity prediction and temporal profile generalization performance, applying to two real-world retweet cascade datasets referencing articles from controversial and traditional media publishers. The \bmh model outperforms the state-of-the-art models and predictive baselines on both datasets and utilizes cascade- and item-level information better than the alternatives. Lastly, we perform a counter-factual analysis where we apply the trained publisher-level \bmh models to a set of article headlines and show that effectiveness of headline writing style (neutral, clickbait, inflammatory) varies across publishers. The \bmh model unveils differences in style effectiveness between controversial and reputable publishers, where we find clickbait to be notably more effective for reputable publishers as opposed to controversial ones, which links to the latter's overuse of clickbait.
\end{abstract}
\section{Introduction}

\begin{figure}
    \centering
    \includegraphics[width=0.81\textwidth]{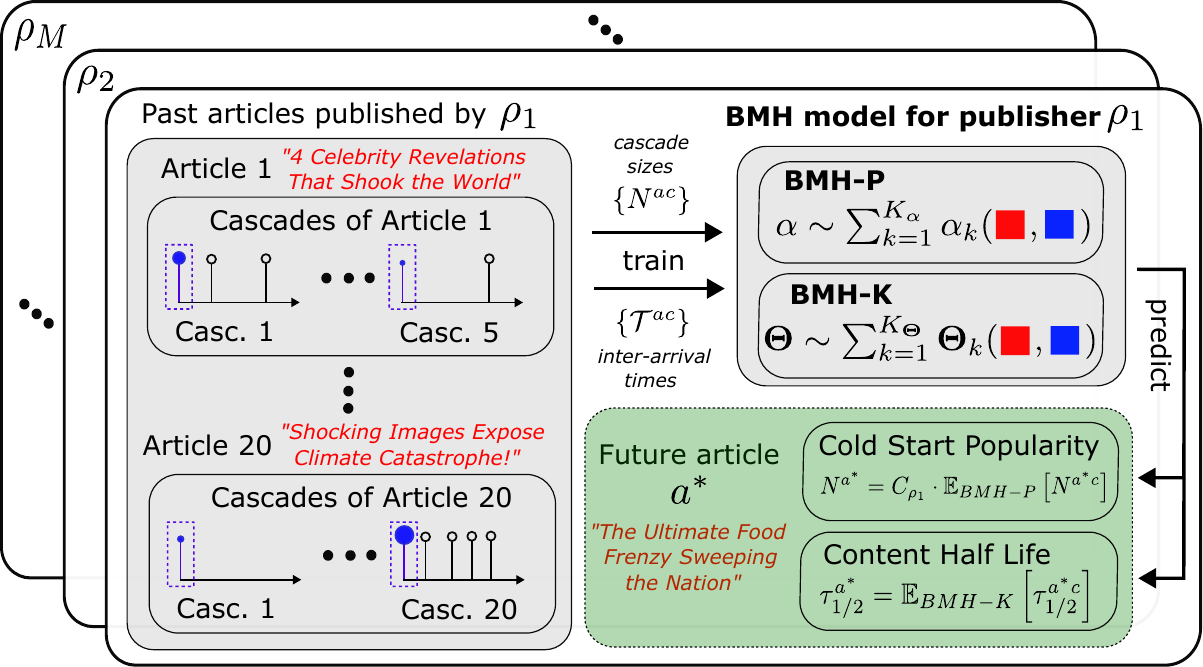}
    \caption{
        An \textit{intuitive plate diagram} for the \bmh model. 
        \textit{Left:} The \bmh model is trained using a historical dataset: a collection of \pioc{$M$ publishers $\{\rho_1, \ldots, \rho_M\}$}, items for each publisher (i.e. articles), and a set of diffusion cascades for each item. 
        \pioc{Each diffusion cascade consists of a timeline of events, here represented by a set of lollipops.}
        \textit{Upper Right:}  The \bmh is a publisher-level model that maps cascade features (shown in \textcolor{blue}{blue color})
        and article features (in \textcolor{red}{red color})
        to a mixture of Hawkes processes. 
        \textit{Lower Right:} The trained \bmh model (with the historical follower count distribution) can be used to infer spread dynamics of future articles based on their headlines.
    } 
    \label{fig:teaser}
\end{figure}

Social media platforms have played an increasingly important role as distribution hubs for content.
In 2023, it was reported that $69\%$ of the U.S. adult population use social media as a news source \cite{Pewresearch}, implying a significant shift in how information is consumed. 
Understanding how content propagates on these platforms -- both the size and speed of dissemination -- is vital since the impact is intrinsically tied to the level of online engagement the content receives.
To command attention in today's digital age, it is not sufficient to craft high-quality content alone, but rather high-quality content that resonates with social media.

The spread of content online is influenced by factors at varying levels. 
At the lowest level, the breadth of a \textit{diffusion cascade}, referring to the sequence of content shares triggered by a user, often hinges on the user's popularity as reflected by their follower count \cite{Bakshy2011}. 
If a highly followed user shares an online item, it reaches a broader audience, increasing the likelihood that it will be shared. 
However, the cascade's growth is not solely dependent on user popularity.
The nature and category of the shared content play crucial roles, as various topics may engage audiences in different ways \cite{Rizoiu2017,Tsagkias2009}. 
For news dissemination, the way an article headline is written, particularly the use of clickbait tactics to create an \textit{information gap} to exploit the audience's curiosity \cite{Zhang2020}, significantly impacts the total attention (i.e. \textit{popularity}) the news article receives. 
Beyond cascade- and item-level factors, the reputation of the online item's source also affects how widely and quickly information spreads \cite{Parikh2019}. 
An article from a reputable source like The New York Times may spread more quickly and be taken more seriously than an article from a controversial, lesser-known blog due to the former's established credibility.
Accurately modeling diffusion cascades of online content requires an approach that jointly considers these factors at different levels.

In this work, we address two open questions related to jointly modeling the influence of the source, item- and cascade-level factors on online content spread.

The first research question examines how these three levels influence the spread of online content. While prior studies have explored the effects of cascade features \cite{Tan2021} and item-level variations \cite{Kong2020}, a comprehensive framework that jointly considers the three levels has yet to be developed. Our first question is: \textbf{Can we build a model for the spread dynamics of online content that accounts for the intertwining influence of its source, the content itself, and cascade-level factors?} 
To tackle this, we propose the Bayesian Mixture Hawkes (\bmh) model, a novel source-level hierarchical mixture model of separable Hawkes processes that models diffusion cascades' size and temporal profile as a function of cascade- and item-level features. The left half of
\cref{fig:teaser} showcases how the source-level \bmh model learns across both the cascade and item levels from a hierarchically structured dataset (i.e., a set of items, cascade groups for each item, and feature sets attached to each).
The \bmh model is capable of learning different classes of Hawkes process dynamics, taking into account the ability of online content to trigger varied responses, from highly popular to largely unnoticed cascades, as well as those that fade quickly or diminish over time. The \bmh learns the influence of feature sets on these classes in two ways: the location of each class in the Hawkes parameter space and the membership probability of each cascade belonging to each class. 
The trained \bmh model can then be used to predict future items' popularity and spread dynamics from the same source (see the right half of \cref{fig:teaser}).
We test the \bmh model on two hierarchical retweet cascade datasets that reference articles from controversial and reputable media publishers \cite{Kong2020}  and on two tasks: cold-start popularity prediction and temporal profile generalization performance. 
We show that the \bmh outperforms the state-of-the-art in item popularity prediction (Dual Mixture Model \cite{Kong2020}), Empirical Bayes approach \cite{Tan2021} and predictive baselines for both tasks and datasets, and that the \bmh model jointly leverages cascade- (i.e., the follower count of the seed user) and article-level (i.e., the article headline embedding vector) information better than the benchmarks.
Furthermore, our model ablation highlights the role of the initiating user in shaping the cascade dynamics related to controversial media, a factor less critical for cascades linked to reputable media.
This distinction mirrors the diverse pathways of online information dissemination: controversial media often circulate within topical social groups \cite{Booth2024,Johns2024}, with the initial endorser serving to validate the content, while for reputable media the publisher's reputation is the most important factor.

Our second open question relates to learning differences in the spread dynamics across news publishers: \textbf{Can we uncover across-publisher differences in how headline writing style (neutral, clickbait, inflammatory) affects published content's popularity and temporal profile?}
We run a counter-factual analysis using the trained publisher-level \bmh models and a labeled set of article headlines \cite{Lee2024} to show the variation of headline style effectiveness across publishers.
We find that the \bmh model is able to capture nuanced publisher behavior, such as the effectiveness of inflammatory headlines for tabloids.
The \bmh model also unveils differences in the success of clickbait between controversial and reputable outlets, linking to existing research on clickbait fatigue and the diminishing relationship between clickbait effectiveness and volume \cite{Zhang2020,Lischka2023}.

\textbf{The main contributions of the work are as follows:}
\begin{enumerate}
    \item The Bayesian Mixture Hawkes (BMH) model\footnote{The Stan/CmdStanPy implementation of the \bmh model is available at \url{https://github.com/behavioral-ds/bayesian-mixture-hawkes/.}\label{footnote:code}}, a novel hierarchical mixture model of the joint influence of cascade- and item-level features on online item spread dynamics. On two news datasets, we show that the \bmh outperforms the state-of-the-art and baselines in cold-start popularity prediction and temporal profile generalization performance.
    \item A counter-factual analysis showing how headline writing style affects published content's spread dynamics. Using the \bmh model we learn the differences in the effectiveness of headlines across publishers and show general trends across controversial and reputable media outlets.
\end{enumerate}

\subsubsection{Related Work.}

In recent years, generative models, and specifically the Hawkes process \cite{Hawkes1971}, have been employed to model online information diffusion given their dual \textit{predictive} and \textit{interpretable} capabilities \cite{Bao2016,Zhao2015,Gomez2011,Ma2016}. 
However, the Hawkes process cannot incorporate feature sets in its base form since it relies only on observed temporal sequences to fit the model parameters. 
Numerous modifications to incorporate feature sets have been proposed to enhance model fit and predictive capabilities. 
A hybrid approach introduced in \cite{Mishra2016} integrates the Hawkes process with a scaling factor trained on cascade-level features to improve retweet cascade size prediction.
The Empirical Bayes (EB) method \cite{Tan2021} utilizes historical retweet sequences to link cascade features and the prior distribution of Hawkes process parameters, leading to better forecasting. 
The parametric Hawkes process \cite{Li2014} models the branching factor, i.e. the expected number of offsprings from a parent event, as a linear combination of event-level features. 
Lastly, the Tweedie-Hawkes process \cite{Li2020} improves on this by combining the Hawkes process with the Tweedie distribution to more realistically model the effect of event-level features on the branching factor. 
The proposed \bmh model is a hierarchical model and can incorporate two levels of feature sets: the cascade- and the item (i.e., cascade-group)-level, which previous work does not cover.

Another relevant area is mixtures of point processes, employed when the data is suspected to be generated from multiple dynamical classes (i.e., parameter sets). 
In \cite{Xu2017}, the Hawkes process was combined with the Dirichlet distribution to model clusters of cascades. 
An online learning framework was introduced in \cite{Ghassemi2022} to fit mixtures of multivariate Hawkes processes to learn the interaction network across a set of actors. \cite{Sharma2019} introduces a generative model for mixtures of more complex point processes by using \pioc{recurrent neural networks}.
Closest to our work is the Dual Mixture Model (DMM) \cite{Kong2020}, a generative model for cascade groups. 
Each cascade is sampled from a mixture of separable Hawkes processes learned jointly with their mixture probabilities. 
To the best of our knowledge, including feature sets into mixture models of point processes has not been explored: the \bmh model solves this by learning the influence of features on the mixture components.

\section{Preliminaries}

We discuss two point process models that form the foundation of the \bmh model.
\cref{prelims:hawkes-process} presents the Hawkes Process (HP) \cite{Hawkes1971}, a temporal point process model that displays self-exciting behavior.
\cref{prelims:dual-mixture-model} introduces the Dual Mixture Model (DMM) \cite{Kong2020}, an approach to jointly model groups of cascades.
An introduction to Bayesian hierarchical modeling, which we employ to model hierarchical data,
is included in Section 1.3 of the Online Appendix \cite{appendix}.

\subsection{Hawkes Process}
\label{prelims:hawkes-process}
The Hawkes process (HP) \cite{Hawkes1971} is a temporal point process widely used to model phenomena that display self-excitation, i.e., the likelihood of an event increases as more events occur. 
The HP is specified using the conditional intensity function $\lambda(t| \mathcal{H})$, the event rate at any time $t$ conditioned on the history  $\mathcal{H} = \{t_j | t_j < t \}$ of past events up to that point, i.e. $\lambda(t | \mathcal{H}) = \mu + \sum_{j=1}^N \alpha \cdot g(t - t_j | \boldsymbol{\Theta})$.
For brevity, we drop the condition on the event history and write $\lambda(t | \mathcal{H})$ as $\lambda(t)$. 
Under this parametrization, a Hawkes process $\mathcal{HP}(\mu, \alpha, \boldsymbol{\Theta} | g)$ is identified with the parameters $\mu$, $\alpha$ and $g(\cdot|\boldsymbol{\Theta}): \mathbb{R}^+ \rightarrow \mathbb{R}^+$. 
The parameter $\mu \geq 0$ \pioc{is the arrival rate of events triggered by external sources}, the branching factor $\alpha \geq 0$ \pioc{is the expected number of offsprings generated by a single parent event which controls the level of self-excitation from previous events}, and the memory kernel $g(\cdot| \boldsymbol{\Theta})$ \pioc{models the temporal decay of influence} of previous events on future events controlled by the parameter set $\boldsymbol{\Theta}$.
In this work, we utilize the power law kernel \pioc{parametrized by $\boldsymbol{\Theta} = \{\theta, d\}$}, given by  $g(t| \theta, d) = \theta \cdot d^\theta \cdot (t+d)^{-(1+\theta)}$.
\pioc{Other common choices for the memory kernel are the exponential kernel $g(t|\theta) = \theta \cdot e^{-\theta t}$ and the Reyleigh kernel $g(t| \theta) = e^{-\frac{1}{2}\theta \cdot t^2}$. We focus on the power law as it has been shown in \cite{Mishra2016} to outperform these alternatives in popularity prediction.}
HP estimation and prediction is discussed in detail in Sec. 1.1 of the Appendix \cite{appendix}.

Given a collection of \textit{complete} cascades $\mathbb{H} = \{\mathcal{H}_i\}$ where each $\mathcal{H}_i$ is completely observed (i.e. terminal time $T_i \rightarrow \infty$),
and assuming no exogenous events (i.e. $\mu = 0$, \pioc{which occurs for instance with Twitter retweet cascades, where all retweets are considered to be spawned by the original tweet}), the HP log-likelihood $\mathcal{L}(\alpha, \boldsymbol{\Theta} | \mathbb{H})$ splits into two log-likelihoods \cite{Kong2020},
\begin{equation}
    \mathcal{L}(\alpha, \boldsymbol{\Theta} | \mathbb{H}) = \mathcal{L}(\alpha | \mathbb{H}) + \mathcal{L}( \boldsymbol{\Theta} | \mathbb{H}),
    \label{eq:separablehawkes}
\end{equation}
\begin{align*}
    \mathcal{L}(\alpha | \mathbb{H}) = \sum_{\mathcal{H}_i \in \mathbb{H}} \log \left[ \alpha^{N_i - 1} e^{-N_i \alpha}\right], &\quad \mathcal{L}(\boldsymbol{\Theta} | \mathbb{H}) = \sum_{\mathcal{H}_i \in \mathbb{H}} \sum_{t_j \in \mathcal{H}_i, j \geq 1} \log \sum_{t_z < t_j} g(t_j - t_z | \boldsymbol{\Theta}),
\end{align*}
where we set $N_i = |\mathcal{H}_i|$. 
Under this case, Hawkes process estimation splits into two independent problems, hence the term \textit{separable Hawkes process}. 
The first problem (popularity estimation) utilizes the cascade sizes $\{N_i\}$ to estimate the branching factor $\alpha$ by maximizing $\mathcal{L}(\alpha | \mathbb{H})$. 
It was shown in \cite{Kong2020} that maximizing $\mathcal{L}(\alpha | \mathbb{H})$ is equivalent to maximizing $\sum_{\mathcal{H}_i \in \mathbb{H}} \log \mathbb{B}(N_i | \alpha)$, where $\mathbb{B}(\cdot | \alpha)$ is the Borel distribution \cite{Borel1942}. The second problem (kernel estimation) uses the interevent-time distribution  $\mathcal{T} = \{t_j - t_z\}_{t_z<t_j, t_j \in \mathcal{H}, \mathcal{H} \in \mathbb{H}}$ to estimate $\boldsymbol{\Theta}$ by maximizing $\mathcal{L}(\boldsymbol{\Theta} | \mathbb{H})$.

\subsection{Dual Mixture Model}
\label{prelims:dual-mixture-model}

Maximizing \cref{eq:separablehawkes} yields the best-fitting Hawkes parameter set $\{\alpha, \boldsymbol{\Theta}\}$.
However, this approach assumes that all cascades stem from a singular parameter set, an assumption which may not hold if there are multiple dynamical classes of cascade behavior.
The Dual Mixture Model (DMM) \cite{Kong2020} was proposed to model a cascade group $\mathbb{H}$ with a mixture of $K$ separable Hawkes processes of different parameter sets to account for different dynamical classes.
Under separability, the DMM splits into two submodels: the Borel mixture model (BMM) for popularity estimation and the kernel mixture model (KMM) for kernel estimation.
The BMM assumes that there exist $K$ popularity classes accounting for the cascade sizes $\{N_i\}$, where the $i^{th}$ class is represented by the branching factor $\alpha_i^*$ with probability $p_i^B$, i.e. $M^B = \{(\alpha_i^*, p_i^B)\}_{i=1}^{K}$.
Similarly, the KMM assumes that there are $K$ kernel classes accounting for the interevent-time distribution $\mathcal{T}$, where the $j^{th}$ class is represented by the kernel parameter set $\boldsymbol{\Theta}^*_j$ with probability $p_j^g$, i.e. $M^g = \{(\boldsymbol{\Theta}_j^*, p_j^g)\}_{i=1}^{K}$. 
The DMM is the Cartesian product of $M^B$ and $M^g$, i.e. $M = \{(\alpha_i^*, \boldsymbol{\Theta}_j^*, p_i^B \cdot p_j^g) | (\alpha_i^*, p_i^B) \in M^B, (\boldsymbol{\Theta}_j^*, p_j^g) \in M^g\}.$ DMM estimation and prediction is discussed in detail in Section 1.2 of the Online Appendix \cite{appendix}.

\section{Bayesian Mixture Hawkes (\bmh) Model}

In this section, we develop the Bayesian Mixture Hawkes (\bmh) model, a hierarchical mixture model of separable Hawkes processes to learn the effect of cascade-level and item-level features on cascade spread dynamics. 
We first describe the dataset structure that the \bmh model is tailored to handle, then discuss the \bmh model's objectives and the approach we adopt to address each. 
We then present the two components of the \bmh: the popularity submodel in \cref{subsection:popularity} and the kernel submodel in \cref{subsection:kernel}. 

\begin{table}[t!]
  \centering
  \caption{Summary of important quantities and notation.}\label{tab:notation}
  \begin{tabularx}{\textwidth}{cXX}
  Parameter & Interpretation & Real-World Mapping \\
  \midrule
  $a/ \mathcal{A}$ & item/s produced by source $\rho$ & news article/s from publisher $\rho$ \\
  $\mathcal{H}^{ac}/ \mathbb{H}^{a}$ & cascade/s related to item $a$ & retweet cascade/s for article $a$ \\
  $\vec{y}^a$ & item-level features of $a$& headline embedding for article $a$ \\
  $N^a$ & item popularity of $a$ & overall tweet count for article $a$\\
  $\vec{x}^{ac}$ & cascade-level features of $\mathcal{H}^{ac}$ & \# followers of $\mathcal{H}^{ac}$ seed user \\
  $N^{ac}$ & cascade size of $\mathcal{H}^{ac}$ & \\
  $\mathcal{T}^{ac}$ & intereevent-time distribution of $\mathcal{H}^{ac}$ & \\
  $(\alpha^{ac}, \boldsymbol{\Theta}^{ac})$ & HP parameter set generating $\mathcal{H}^{ac}$ & \\
  $\tau_{1/2}^{ac}$ & diffusion half-life of $\mathcal{H}^{ac}$ & \\
  $K_{\alpha}/ K_{\boldsymbol{\Theta}}$ & \# of \bmhp/-K classes  & \\
  $z^{ac}_{\alpha,k}/ z^{ac}_{\boldsymbol{\Theta},k}$ & class $k$ membership probability& \\
  $\delta_{\alpha,k}/ \delta_{\theta,k}$ & baseline $\text{logit}(\alpha)$, $\text{log}(\theta)$  for class $k$ & \\
  $\delta_{z_{\alpha,k}}/ \delta_{z_{\boldsymbol{\Theta},k}}$ & baseline class $k$ mem. probability & \\
  $\vec{\gamma}_{\alpha,k}/ \vec{\gamma}_{\theta,k}$ & effect of $\vec{y}^{a}$ on class $k$ center  & \\
  $\vec{\gamma}_{z_{\alpha,k}}/ \vec{\gamma}_{z_{\boldsymbol{\Theta},k}}$ & effect of $\vec{y}^{a}$ on class $k$ mem. prob.& \\
  $\vec{\beta}_{\alpha,k}/ \vec{\beta}_{\theta,k}$ & effect of $\vec{x}^{ac}$ on class $k$ center& \\
  $\vec{\beta}_{z_{\alpha,k}}/ \vec{\beta}_{z_{\boldsymbol{\Theta},k}}$ & effect of $\vec{x}^{ac}$ on class $k$ mem. prob.& \\
 \bottomrule
  \end{tabularx}
\end{table}

Assume that we are given the following dataset. First, we have a collection of items, denoted as $\mathcal{A}$, from a shared source $\rho$, where each item $a \in \mathcal{A}$ is characterized by the feature vector $\vec{y}^{a} \in \mathbb{R}^{N_y}$. If $\rho$ is a news publisher, then $\mathcal{A}$ can represent a collection of news articles and $\vec{y}^{a}$ the embedding vector for article $a$'s headline. Second, we have a set of complete cascades $\mathbb{H}^{a}$ for each item $a \in \mathcal{A}$, where cascade $\mathcal{H}^{ac} \in \mathbb{H}^a$ has size $N^{ac}$, interevent distribution $\mathcal{T}^{ac}$, and is described by the feature vector $\vec{x}^{ac} \in \mathbb{R}^{N_x}$. \pioc{In our news example, $\mathbb{H}^a$ can represent discussions on Twitter related to article $a$, which we obtain by collecting all retweet cascades initiated with a tweet linking article $a$'s URL. The feature vector} $\vec{x}^{ac}$ can be taken as the follower count of the cascade's initiator.

We model the generative process of $\mathcal{H}^{ac}$ using a separable power-law HP with parameter set $(\alpha^{ac}, \boldsymbol{\Theta}^{ac})$, i.e. $\mathcal{H}^{ac} \sim \mathcal{HP}(\alpha^{ac}, \boldsymbol{\Theta}^{ac} | g)$. 
We construct the \bmh as a model for $(\alpha^{ac}, \boldsymbol{\Theta}^{ac})$ with three goals: (1) jointly learn across the item set $\mathcal{A}$, (2) learn the relationship between $\vec{y}^{a}$ and $({\alpha^{ac}, \boldsymbol{\Theta}^{ac}})$, and (3) learn the link between $\vec{x}^{ac}$ and the same parameters.
We handle goal (1) by using a two-level Bayesian hierarchical model to jointly fit across each item $a \in \mathcal{A}$ and to tie together cascade- and item-level information.
For goals (2) and (3), we consider a mixture of separable HPs with $K_{\alpha}$ classes for $\alpha^{ac}$ and $K_{\boldsymbol{\Theta}}$ classes for $\boldsymbol{\Theta}^{ac}$. We learn the influence of $\vec{y}^{a}$ and $\vec{x}^{ac}$ on $\{\alpha^{ac}, \boldsymbol{\Theta}^{ac}\}$ through the centers and membership probabilities of the $K_{\alpha}$ popularity classes and $K_{\boldsymbol{\Theta}}$ kernel classes. 

Due to the separability of the underlying HP, the \bmh divides into two independent models: (1) \bmhp, the \textit{popularity} submodel for $\alpha^{ac}$, and (2) \bmhk, the \textit{kernel} submodel for $\boldsymbol{\Theta}^{ac}$. 
\cref{tab:notation} lists the notation for important variables in the \bmh and the mapping to real-world quantities in the datasets in \cref{sec:predictive}. 

\subsection{\bmhp, the Popularity Submodel}
\label{subsection:popularity}

\begin{figure}[t!]
    \centering
    \includegraphics[width=0.75\textwidth]{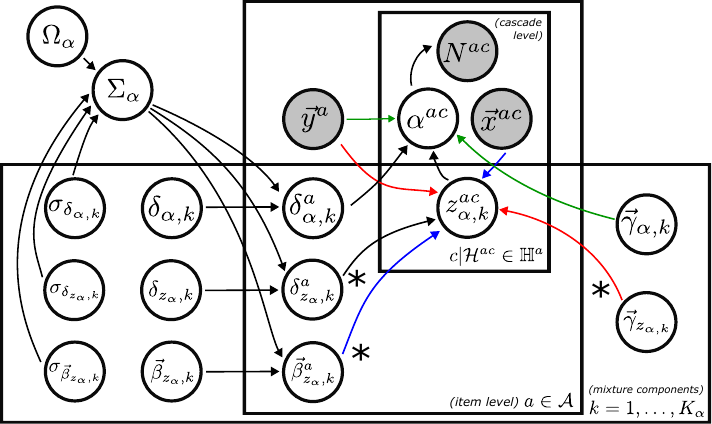}
    \caption{Plate diagram of the \bmhp model. 
    Shaded nodes are observables while empty nodes are latent variables.
    Paired colored edges indicate source nodes appearing as a product in the target node. 
    For instance, the \textcolor{olive}{green} edges indicate that $\protect\vec{\gamma}_{\alpha,k}$ and $\protect\vec{y}^a$ appear as $\protect\vec{\gamma}_{\alpha,k} \cdot \protect\vec{y}^a$ in the expression for $\alpha^{ac}$ in \cref{eq:logit_a}. 
    The same concept holds for the \textcolor{blue}{blue} and \textcolor{red}{red} edges.
    Edges marked with * indicate dependence of the target node on the source node indexed with $k$ and the entire set $\{1, \cdots, K_{\alpha}\}$.
    For instance, in \cref{eq:z_ac} $z^{ac}_{{\alpha},k}$ depends on $\protect\vec{\beta}^a_{z_{{\alpha}, k}}$ (see the numerator) and $\protect\vec{\beta}^a_{z_{{\alpha}, k'}}$ for $k' \in \{1, \cdots, K_{\alpha}\}$ (see the denominator).
    } 
    \label{fig:platealpha}
\end{figure}

The branching factor $\alpha^{ac}$ is modeled as the mixture random variable
\begin{equation}
    \text{logit}(\alpha^{ac}) = \delta^{a}_{\alpha, k} + \textcolor{olive}{\vec{\gamma}_{\alpha,k} \cdot \vec{y}^a}, \label{eq:logit_a}
\end{equation}
with membership probability $z^{ac}_{\alpha,k}$ ($k = 1, \ldots, K_\alpha$),
\begin{equation}
    z^{ac}_{\alpha,k} =
        \frac{\text{exp}(\delta^{a}_{z_{\alpha, k}} + \textcolor{blue}{\vec{\beta}^a_{z_{\alpha, k}} \cdot \vec{x}^{ac}} + \textcolor{red}{\vec{\gamma}_{z_{\alpha, k}} \cdot \vec{y}^a})}{\sum_{k'=1}^{K_\alpha}\text{exp}(\delta^{a}_{z_{\alpha, k'}} + \textcolor{blue}{\vec{\beta}^a_{z_{\alpha, k'}} \cdot \vec{x}^{ac}} + \textcolor{red}{\vec{\gamma}_{z_{\alpha, k'}} \cdot \vec{y}^a})}. \label{eq:z_ac}
\end{equation}
The intercept $\delta^{a}_{\alpha, k}$ in \cref{eq:logit_a} sets the centering of $\text{logit}(\alpha^{ac})$ for popularity class $k$.
In \cref{eq:z_ac}, we designate $k=1$ as the reference class (i.e. $\delta^{a}_{z_{\alpha, 1}} = \vec{\beta}^a_{z_{\alpha, 1}} = \vec{\gamma}_{z_{\alpha, 1}} =0$); parameters for $k>1$ control deviation from class $k=1$. The intercept $\delta^{a}_{z_{\alpha, k}}$ controls the baseline proportion of class $k$'. 
The influence of item features on $\text{logit}(\alpha^{ac})$ and class $k$ membership are estimated by $\vec{\gamma}_{{\alpha, k}}$ and $\vec{\gamma}_{{z_{\alpha, k}}}$, respectively, while the influence of cascade features on class $k$ membership is estimated by $\vec{\beta}^a_{{z_{\alpha,k}}}$. Note that $\vec{\gamma}_{{\alpha,k}},\vec{\gamma}_{{z_{\alpha, k}}}$ are shared across $\mathcal{A}$ while $\vec{\beta}^a_{{z_{\alpha,k}}}$ is estimated per $a$. 

For brevity, we collect the parameter vector specific to item $a$ as $\vec{p}_{\alpha}^a = [\delta^{a}_{\alpha, 1}, \ldots, \delta^{a}_{\alpha, K_\alpha}, \delta^{a}_{z_{\alpha, 2}}, \ldots, \delta^{a}_{z_{\alpha, K_\alpha}}, \vec{\beta}^a_{z_{\alpha,2}}, \ldots, \vec{\beta}^a_{z_{\alpha,K_\alpha}}]^\intercal.$
We link item $a$ with $\mathcal{A}$ by assuming that $\vec{p}_{\alpha}^a$ is drawn from a source-level multivariate normal (MVN) distribution with mean $\vec{p}_{\alpha}$ and covariance matrix $\Sigma_{\alpha}$,
\begin{align}
    \vec{p}_{\alpha}^a \sim \text{MVN}\left(\vec{p}_{\alpha}, \Sigma_{\alpha}\right), \quad \Sigma_{\alpha} &= \text{D}_\alpha \cdot \Omega_\alpha \cdot \text{D}_\alpha, \quad \text{D}_\alpha = \text{diag}(\sigma_{\vec{p}_\alpha}) \label{eq:alpha_artistlevel},
\end{align}
where $\Omega_\alpha$ is a correlation matrix and $\sigma_{\vec{p}_\alpha}$ is a vector of standard deviations corresponding to $\vec{p}_\alpha$. 

The plate diagram for the \bmhp model is shown in \cref{fig:platealpha}. Variable pairs that appear as a product term are colored \textcolor{olive}{green}, \textcolor{red}{red} and \textcolor{blue}{blue} in \cref{eq:logit_a,eq:z_ac}, visualized in \cref{fig:platealpha} as source nodes with \textcolor{olive}{green}, \textcolor{red}{red} and \textcolor{blue}{blue} edges.

\subsubsection{Inference and Prediction.} Let $\mathcal{P}_\alpha$ be the parameter set for the \bmhp model. From the set of cascade sizes $\{N_{ac}\}_{\mathcal{H}^{ac} \in \mathbb{H}^a, a \in \mathcal{A}}$, we estimate the posterior distribution $\mathbb{P}(\mathcal{P}_{\alpha} | \{N_{ac}\}_{ac} ) \propto \exp(\mathcal{L}(\mathcal{P}_{\alpha} |  \{N_{ac}\}_{ac} )) \cdot \mathbb{P}(\mathcal{P}_{\alpha})$,
where $\mathbb{P}(\mathcal{P}_{\alpha})$ is the prior for $\mathcal{P}_{\alpha}$ and  $\mathcal{L}(\mathcal{P}_{\alpha} |  \{N_{ac}\}_{ac} )$ is the log-likelihood of $\mathcal{P}_\alpha$ given the cascade sizes (derived in Section 2.2 of the Online Appendix \cite{appendix}).
\pioc{Informative priors have to be set on $\{\delta_{\alpha,k},\delta_{z_{\alpha,k}}\}$ to identify the $K_\alpha$ classes in the $\alpha$ parameter space. $\delta_{\alpha,k}$ and $\delta_{z_{\alpha,k}}$ identify the center and baseline proportion of the $k^{th}$ class, respectively.} Weakly informative priors are set for the other parameters in $\mathcal{P}_\alpha$. \pioc{We implement\footref{footnote:code} the \bmhp model in Stan\cite{Carpenter2017}, which uses the No-U-Turn Sampler (NUTS), a Hamiltonial Monte Carlo technique, to sample the posterior distribution $\mathbb{P}(\mathcal{P}_{\alpha} | \{N_{ac}\}_{ac} )$. 
We use CmdStanPy \cite{cmdstanpy} to run Stan code through Python.}

Using the average cascade count for items in $\mathcal{A}$, denoted as $\hat{C}_\rho$, and the empirical distribution of the cascade feature vector $\vec{x}^{ac}$, denoted as $\hat{f}_\rho({x})$, the fitted \bmhp model can be used to estimate the cold-start popularity $\hat{N}^{a^*}$ of an out-of-sample item $a^*$ with feature vector $\vec{y}^{a^*}$:
\begin{equation}
    \hat{N}^{a^*} \approx \hat{C}_\rho \cdot \sum_{x=0}^\infty \sum_{k=1}^{K_\alpha}  z_{\alpha, k}^{a^*,c} \cdot \left[ 1 + \text{exp}\left(\delta^{a^*}_{\alpha, k} + \vec{\gamma}_{\alpha,k} \cdot \vec{y}^{a^*} \right) \right] \cdot \hat{f}_\rho(x), \label{eq:coldstartbayes}
\end{equation}
where we assume that $\vec{x}^{ac} = x \in \mathbb{N}$ (see Section 2.2 of the Appendix \cite{appendix}).

\subsection{\bmhk, the Kernel Submodel}
\label{subsection:kernel}

\begin{figure}[t!]
    \centering
    \includegraphics[width=0.82\textwidth]{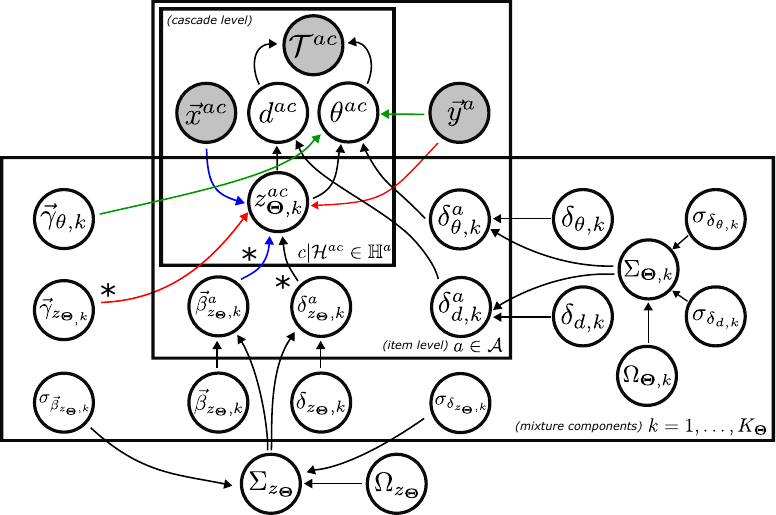}
    \caption{Plate diagram of the \bmhk model. Shaded nodes are observables while empty nodes are latent variables. 
    Paired colored edges indicate source nodes appearing as a product in the target node.
    For instance, the \textcolor{olive}{green} edges indicate that $\protect\vec{\gamma}_{\theta,k}$ and $\protect\vec{y}^a$ appear as the product $\protect\vec{\gamma}_{\theta,k} \cdot \protect\vec{y}^a$ in the expression for $\theta^{ac}$ in \cref{eq:logit_t1}. The same concept holds for the \textcolor{blue}{blue} and \textcolor{red}{red} edges. 
    Edges marked with * indicate dependence of the target node on the source node indexed with $k$ and the entire set $\{1, \cdots, K_{\boldsymbol{\Theta}}\}$.
    For instance, in \cref{eq:z_ac_theta} $z^{ac}_{{\boldsymbol{\Theta}},k}$ depends on $\protect\vec{\beta}^a_{z_{{\boldsymbol{\Theta}}, k}}$ (see the numerator) and $\protect\vec{\beta}^a_{z_{{\boldsymbol{\Theta}}, k'}}$ for $k' \in \{1, \cdots, K_{\boldsymbol{\Theta}}\}$ (see the denominator).
    } \label{fig:platetheta}
\end{figure}

Under the power-law, the kernel parameter set generating $\mathcal{H}^{ac}$ is $\boldsymbol{\Theta}^{ac} = [\theta^{ac}, d^{ac}]^{\intercal}$. 
We model $\boldsymbol{\Theta}^{ac}$ as a pair of mixture random variables taking the value 
\begin{align}
    \text{log}(\theta^{ac})= \delta^{a}_{\theta, k} + \textcolor{olive}{\vec{\gamma}_{\theta,k} \cdot \vec{y}^a}, &\quad \text{log}(d^{ac})= \delta^{a}_{d, k}  \label{eq:logit_t1}
\end{align}
with probability $z^{ac}_{\boldsymbol{\Theta},k}$ ($k = 1, \ldots, K_{\boldsymbol{\Theta}}$), where
\begin{equation}
    z^{ac}_{{\boldsymbol{\Theta}},k} =
        \frac{\text{exp}(\delta^{a}_{z_{{\boldsymbol{\Theta}}, k}} + \textcolor{blue}{\vec{\beta}^a_{z_{{\boldsymbol{\Theta}}, k}} \cdot \vec{x}^{ac}} + \textcolor{red}{\vec{\gamma}_{z_{{\boldsymbol{\Theta}}, k}} \cdot \vec{y}^a})}{\sum_{k'=1}^{K_{\boldsymbol{\Theta}}}\text{exp}(\delta^{a}_{z_{{\boldsymbol{\Theta}}, k'}} + \textcolor{blue}{\vec{\beta}^a_{z_{{\boldsymbol{\Theta}}, k'}} \cdot \vec{x}^{ac}} + \textcolor{red}{\vec{\gamma}_{z_{{\boldsymbol{\Theta}}, k'}} \cdot \vec{y}^a})}.
         \label{eq:z_ac_theta}
\end{equation}

In \cref{eq:z_ac_theta} we designate $k=1$ as the reference class (i.e. $\delta^{a}_{z_{{\boldsymbol{\Theta}}, 1}} = \vec{\beta}^a_{z_{{\boldsymbol{\Theta}}, 1}} = \vec{\gamma}_{z_{{\boldsymbol{\Theta}}, 1}} =0)$.

Let $\vec{p}_{\boldsymbol{\Theta}, k}^a = [\delta^{a}_{\theta, k}, \delta^{a}_{d, k}]^\intercal$ and $\vec{p}_{z_{\boldsymbol{\Theta}}}^a = \left[\delta_{z_{\boldsymbol{\Theta}, 2},}^a, \ldots, \delta_{z_{\boldsymbol{\Theta}, K_{\boldsymbol{\Theta}}}}^a, \vec{\beta}^a_{z_{\boldsymbol{\Theta},2}}, \ldots, \vec{\beta}^a_{z_{\boldsymbol{\Theta},K_{\boldsymbol{\Theta}}}} \right]^\intercal$. The complexity of estimating two parameters (i.e. $\theta^{ac}, d^{ac}$) makes it challenging to estimate a joint source-level MVN distribution as we did for \bmhp. To simplify, we assume independence of $(\delta^{a}_{\theta, k}, \delta^{a}_{d, k})$ across classes. 
For each kernel class $k$, we assume $\vec{p}_{\boldsymbol{\Theta},k}$ is drawn from a source-level MVN distribution with mean $\vec{p}_{\boldsymbol{\Theta}, k} = [\delta_{\theta, k}, \delta_{d, k}]^\intercal$ and covariance matrix $\Sigma_{\boldsymbol{\Theta}, k}$.
Lastly, we assume $\vec{p}_{z_{\boldsymbol{\Theta}}}^a$ is drawn from an MVN distribution with mean $\vec{p}_{z_{\boldsymbol{\Theta}}}$ and covariance matrix $\Sigma_{z_{\boldsymbol{\Theta}}}$.

\begin{align}
    \vec{p}_{{\boldsymbol{\Theta}},k}^a \sim \text{MVN}\left(\vec{p}_{{\boldsymbol{\Theta},k}}, \Sigma_{{\boldsymbol{\Theta},k}}\right) , \quad \Sigma_{{\boldsymbol{\Theta},k}} &= \text{D}_{\boldsymbol{\Theta},k} \cdot \Omega_{\boldsymbol{\Theta},k} \cdot \text{D}_{\boldsymbol{\Theta},k}, \quad \text{D}_{\boldsymbol{\Theta},k} = \text{diag}(\sigma_{\vec{p}_{\boldsymbol{\Theta},k}})  \nonumber \\    
     \vec{p}_{z_{\boldsymbol{\Theta}}}^a \sim \text{MVN}\left(\vec{p}_{z_{\boldsymbol{\Theta}}}, \Sigma_{z_{\boldsymbol{\Theta}}}\right), \quad \Sigma_{z_{\boldsymbol{\Theta}}} &= \text{D}_{z_{\boldsymbol{\Theta}}} \cdot \Omega_{z_{\boldsymbol{\Theta}}} \cdot \text{D}_{z_{\boldsymbol{\Theta}}}, \quad \text{D}_{z_{\boldsymbol{\Theta}}} = \text{diag}(\sigma_{z_{\boldsymbol{\Theta}}}),  \nonumber
\end{align}
where $\sigma_{\vec{p}_{\boldsymbol{\Theta},k}}, \sigma_{\vec{p}_{z_{\boldsymbol{\Theta}}}}$ are standard deviation vectors and $\Omega_{\boldsymbol{\Theta},k}, \Omega_{z_{\boldsymbol{\Theta}}}$ are correlation matrices. 

The plate diagram for the \bmhk model is shown in \cref{fig:platetheta}. Variable pairs that appear as a product term are colored \textcolor{olive}{green}, \textcolor{red}{red} and \textcolor{blue}{blue} in \cref{eq:logit_t1,eq:z_ac_theta}, visualized in \cref{fig:platetheta} as source nodes with \textcolor{olive}{green}, \textcolor{red}{red} and \textcolor{blue}{blue} edges.

\subsubsection{Inference and Prediction.} Let $\mathcal{P}_{\boldsymbol{\Theta}}$ be the parameter set for the \bmhk model.  From the interevent-time distributions $\{\mathcal{T}^{ac}\}_{ac}$, we estimate the posterior distribution $\mathbb{P}(\mathcal{P}_{\boldsymbol{\Theta}} | \mathcal{T}^{ac}) \propto \exp(\mathcal{L} (\mathcal{P}_{\boldsymbol{\Theta}} | \{\mathcal{T}^{ac}\}_{ac}) \cdot \mathbb{P}(\mathcal{P}_{\boldsymbol{\Theta}})$.
The log-likelihood of $\mathcal{P}_{\boldsymbol{\Theta}}$ given $\{\mathcal{T}^{ac}\}_{ac}$ is derived in Section 2.3 of Online Appendix \cite{appendix}.
\pioc{Informative priors have to be set on $\{\delta_{\theta, k}, \delta_{d, k}, \delta_{z_{{\boldsymbol{\Theta}}, k}}\}$ to identify the $K_{\boldsymbol{\Theta}}$ classes in the $(\theta, d)$ parameter space. $(\delta_{\theta, k}, \delta_{d, k})$ and $\delta_{z_{{\boldsymbol{\Theta}}, k}}$ identify the center and baseline proportion of the $k^{th}$ class, respectively.} Weakly informative priors are set for the other parameters in $\mathcal{P}_{\boldsymbol{\Theta}}$. \pioc{Similar to the \bmhp model, we implementt\footref{footnote:code} the \bmhk model in Stan and CmdStanPy to sample from the posterior distribution $\mathbb{P}(\mathcal{P}_{\boldsymbol{\Theta}} | \mathcal{T}^{ac})$.}

The \bmhk model predicts the half-life $\hat{\tau}_{1/2}^{a^*}$ of an out-of-sample item $a^*$ as (see Section 2.3 of the Online Appendix \cite{appendix}),
\begin{align}
    \hat{\tau}_{1/2}^{a^*}
    &\approx \sum_{x=0}^\infty \sum_{k=1}^{K_{\boldsymbol{\Theta}}}  z_{\boldsymbol{\Theta}, k}^{a^*,c} \cdot e^{\delta^{a^*}_{d, k}} \cdot \left[2^{\exp{(\delta^{a^*}_{{{\theta}}, k} + \vec{\gamma}_{{{\theta}},k} \cdot \vec{y}^{a^*}})} - 1 \right] \cdot \hat{f}_\rho(x). \label{eq:halflifebayes}
\end{align}

\section{Predictive Evaluation}
\label{sec:predictive}

In this section, we introduce two evaluation datasets (\cref{subsection:datasets}) and assess the \bmh model's performance on two tasks: cold-start popularity prediction (\cref{subsection:coldstart})
and temporal profile generalization performance (\cref{subsection:temporal}), i.e. evaluating the likelihood of the interevent distribution of future cascades.

\begin{table}[t!]
    \caption{Statistics of the predictive evaluation datasets.}\label{tab:predictive}
    \centering
    \begin{tabularx}{\textwidth}{XXXXX}
    \hline
    & $CNIX-Fit$ & $CNIX-Test$ & $RNIX-Fit$ & $RNIX-Test$\\
    \hline
    \#publishers & 41 & 41 & 28 & 28\\
    \#articles & 72,009 & 40,506 & 2,682 & 18,116\\
    \#cascades & 4,620,509 & 1,874,729 & 244,596 & 460,504\\
    \#tweets & 42,546,067 & 18,235,185 & 1,573,909 & 5,139,967\\
    \hline
    \end{tabularx}
    \end{table}
\subsection{Datasets}
\label{subsection:datasets}

We use two datasets from \cite{Kong2020} for predictive evaluation, consisting of collections of Twitter retweet cascades that link articles from online news sources. 
The Controversial News Index (\textit{CNIX}) dataset consists of retweet cascades mentioning articles from 41 online news publishers known for controversial content, such as \url{https://www.breitbart.com/}. Conversely, the Reputable News Index (\textit{RNIX}) follows the same structure as the $CNIX$ dataset but gathers cascades linked to articles from 28 reputable publishers, such as \url{https://www.news.com.au/}. The tweets for both datasets were collected by the QUT Digital Media Research Centre by retrospectively querying the Twitter search endpoint for URL mentions of the articles between June 30, 2017 and Dec 31, 2019. In \cref{tab:notation} we link quantities in these datasets with variables in the \bmh model.

Both \textit{CNIX} and \textit{RNIX} are temporally split into \textit{Fit} (i.e. training) and \textit{Test} (i.e. evaluation) datasets. The first contains tweets published from Jun 30, 2017 to Jan 1, 2019, while the second contains tweets from Feb 1, 2019 to Dec 31, 2019. 
A one-month gap between \textit{Fit} and \textit{Test} ensures that cascades in the training data are finished before the test period. \cref{tab:predictive} shows summary statistics.

We use the standardized 32-D embedding of $a$'s headline (i.e. PCA-reduced, \textit{all-MiniLM-L6-v2} \cite{Remiers2019}) as our article feature vector $\vec{y}^{a}$, and the standardized log-follower count of the cascade's seed user as the cascade feature vector $\vec{x}^{ac}$.

\subsection{Cold-Start Popularity Prediction}
\label{subsection:coldstart}

Our first task is evaluating the ability of the \bmhp model to predict cold-start popularity of unpublished content. With publisher $\rho$'s trained \bmhp model, we predict the future popularity ${N}^{a^*}$ of an out-of-sample article $a^*$ with \cref{eq:coldstartbayes}.
\pioc{To guide the selection of the number of mixture components $K_\alpha$, we fit the BMM to each publisher in $RNIX$. 
We observe that the BMM-fitted $\{\alpha^a_i\}$ distribution is bimodal, corresponding to clusters of popular and unpopular cascades. 
See Section 3.1 of the Appendix \cite{appendix} for full details. 
Using this result, we fit a \bmhp model for each publisher in $CNIX$ and $RNIX$ in Stan with the hyperparameter $K_\alpha=2$. 
The full set of priors for the \bmhp model is listed in Section 3.2 of the Appendix \cite{appendix}. 
Note that we use a Laplace prior on $\vec{\gamma}_{\alpha,1}, \vec{\gamma}_{\alpha,2}, \vec{\gamma}_{z_{\alpha,2}}$ to impose regularization given the high dimensionality of the article feature vector ($|\vec{y}^{a}| = 32$) we consider}.

To evaluate the predictive power of $\vec{x}^{ac}$ and $\vec{y}^{a}$, apart from the full model as developed in \cref{subsection:popularity} (which we call $\alpha(\vec{y}^{a}) + z(\vec{x}^{ac},\vec{y}^{a})$) we fit three simpler variants of \bmhp: (1) $\alpha(\vec{y}^{a})+z(\vec{y}^{a})$, where we set $\vec{x}^{ac} = 0$ in \cref{eq:z_ac}; (2) $\alpha(\varnothing)+z(\vec{y}^{a})$, where set $\vec{x}^{ac} = 0$ in \cref{eq:z_ac} and $\vec{y}^{ac} = 0$ in \cref{eq:logit_a}; and (3) $\alpha(\varnothing)+z(\varnothing)$, where we set $\vec{x}^{ac} = 0$ in \cref{eq:z_ac} and $\vec{y}^{ac} = 0$ in \cref{eq:logit_a,eq:z_ac}.

We compare the performance of the \bmhp model to three approaches: (1) the DMM \cite{Kong2020}, (2) the empirical Bayes (EB) approach \cite{Tan2021}, and (3) feature-based cascade-size (CR) regression models (i.e. a neural network with one hidden layer of 100 nodes) built using scikit-learn \cite{Pedregosa2011}. For EB and CR, we fit two variants: one using only article features (i.e. EB(y) and CR(y)) and another using both cascade and article features (i.e. EB(x,y) and CR(x,y)). 
We report the Average Relative Error ($ARE$) over the set of articles in the $Test$ datasets. Let ${N}^a$ and $\hat{N}^a$ be the actual and predicted popularity of article $a$, then $ARE(a) = \frac{|\hat{N}^{a} - N^a|}{N^a}$.

\subsubsection{Results.}
In the top half of \cref{tab:ablation}, we summarize cold-start popularity prediction performance of the model variants for $CNIX/RNIX$. 
In both datasets the variants with only article-level features $\vec{y}^a$ and without the cascade-level features $\vec{x}^{ac}$ show minimal performance gain ($RNIX$) or even worse performance ($CNIX$) over the no-feature $\alpha(\varnothing)+z(\varnothing)$ model. 
The full model $\alpha(\vec{y}^{a}) + z(\vec{x}^{ac},\vec{y}^{a})$ significantly outperforms each simpler variant, highlighting the importance of the seed user's popularity as a predictor of final popularity \cite{Bakshy2011}.

We compare the performance of the best-performing \bmhp model with the benchmarks in the top row of \cref{fig:predictive}(a) and \cref{fig:predictive}(b). 
We can see that the \bmhp model outperforms each benchmark based on median performance. 
We note that in each task, the benchmarks that only have article features ($CR(y)$ and $EB(y)$) outperform the corresponding benchmarks that also include cascade features ($CR(x,y)$ and $EB(x,y)$).
However, our ablation results show that the best-performing \bmhp model includes both the cascade and article features.
This implies that the added structure of the \bmhp model jointly leverages the article- and cascade-level information better than the benchmarks.

\begin{table}[t!]
  \centering
  \caption{Popularity prediction and generalization results. We show the median ($25^{th}, 75^{th}$ quantiles) for \bmh with different feature components removed. \pioc{Lower ARE/NLL mean better performance.} The best score across variants is in bold.}\label{tab:ablation}
  \begin{tabularx}{\textwidth}{XXX}
  \toprule
   Popularity (ARE) & $CNIX$ & $RNIX$\\
  \midrule
  $\alpha(\varnothing)+z(\varnothing)$ & 0.707 (0.334, 1.513) & 0.644 (0.335, 0.921) \\
  $\alpha(\varnothing)+z(\vec{y}^{a})$ & 0.708 (0.336, 1.497) & 0.666 (0.339, 1.033) \\
  $\alpha(\vec{y}^{a})+z(\vec{y}^{a})$ & 0.738 (0.370, 1.316) & 0.643 (0.325, 0.953) \\
  $\alpha(\vec{y}^{a})+z(\vec{x}^{ac},\vec{y}^{a})$ & \textbf{0.646} (0.313, 0.935) & \textbf{0.635} (0.342, 0.932)
  \\ \bottomrule\toprule
  Generalization (NLL) & $CNIX$ & $RNIX$\\
  \hline
  $\theta(\varnothing)+z(\varnothing)$ & -3.841 (-5.293, -2.717) & -2.564 (-3.231, -2.031)\\
  $\theta(\varnothing)+z(\vec{y}^{a})$ & -3.782 (-4.873, -2.683) & -2.550 (-3.226, -1.988) \\
  $\theta(\vec{y}^{a})+z(\vec{y}^{a})$ & -3.649 (-4.816, -2.617) & \textbf{-2.689} (-3.492, -2.117) \\
  $\theta(\vec{y}^{a})+z(\vec{x}^{ac},\vec{y}^{a})$ & \textbf{-4.013} (-5.766, -2.714) & -2.645 (-3.450, -2.063)\\
  \bottomrule
  \end{tabularx}
\end{table}

\subsection{Temporal Profile Generalization Performance}
\label{subsection:temporal}

Our second task is evaluating the performance of the \bmhk model in capturing the inter-arrival distribution of future cascades of unpublished articles. Given publisher $\rho$'s trained \bmhk model, we calculate the log-likelihood $\mathcal{L} (\mathcal{P}_{\boldsymbol{\Theta}} | \{\mathcal{T}^{a^*c}\})$ of the inter-arrival distribution $\{\mathcal{T}^{a^*c}\}$ of an unpublished article $a^*$.

\pioc{To guide the selection of the number of mixture components $K_{\boldsymbol{\Theta}}$, we fit the KMM to each publisher in $RNIX$. We observe that the KMM-fitted $\{\theta^a_i, d^a_I\}$ distribution is trimodal, corresponding to clusters of usual, slow- and fast-diffusing cascades cascades. See Section 3.1 of the Appendix \cite{appendix} for full details. Using this result, we fit a \bmhk model for each publisher in $CNIX$ and $RNIX$ in Stan with the hyperparameter $K_{\boldsymbol{\Theta}}=3$.
The full set of priors for the \bmhk model is listed in Section 3.3 of the Appendix \cite{appendix}. 
Note that we use a Laplace prior on $\vec{\gamma}_{{\boldsymbol{\Theta},2}}, \vec{\gamma}_{{\boldsymbol{\Theta},3}}, \vec{\gamma}_{z_{\boldsymbol{\Theta},2}}, \vec{\gamma}_{z_{{\boldsymbol{\Theta},3}}}$ to impose regularization given the high dimensionality of the article feature vector ($|\vec{y}^{a}| = 32$) we consider}.

In addition to the full \bmhk model developed in \cref{subsection:kernel} (which we call $\theta(\vec{y}^{a}) + z(\vec{x}^{ac},\vec{y}^{a})$) we fit three progressively simpler variants analogous to the ablation for the \bmhp model: $\theta(\vec{y}^{a})+z(\vec{y}^{a})$, $\theta(\varnothing)+z(\vec{y}^{a})$, and $\theta(\varnothing)+z(\varnothing)$.
To evaluate performance, we calculate the loglikelihood $\mathcal{L} (\mathcal{P}_{\boldsymbol{\Theta}} | \{\mathcal{T}^{ac}\})$ of inter-arrival times $\{\mathcal{T}^{ac}\}_{a \in \mathcal{A}}$ over articles in the $Test$ datasets. Since we are evaluating on likelihood, we use generative models as benchmarks: the DMM, EB(y), EB(x,y), and publisher-level joint HP (see Sec. 1.1 of the Appendix \cite{appendix}).

\subsubsection{Results.} 
\begin{figure}[t!]
    \centering
    \begin{subfigure}[t]{0.43\textwidth}
        \centering
        \includegraphics[width=\textwidth]{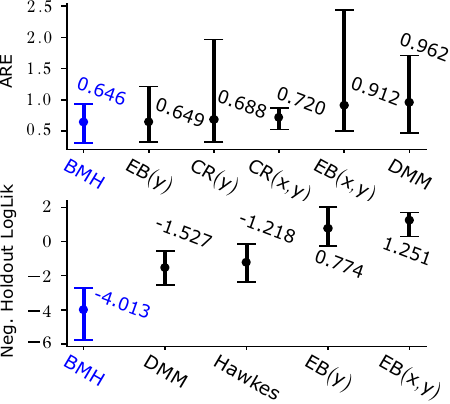}
        \caption{}
    \end{subfigure}
    \hfill
    \begin{subfigure}[t]{0.43\textwidth}
        \centering
        \includegraphics[width=\textwidth]{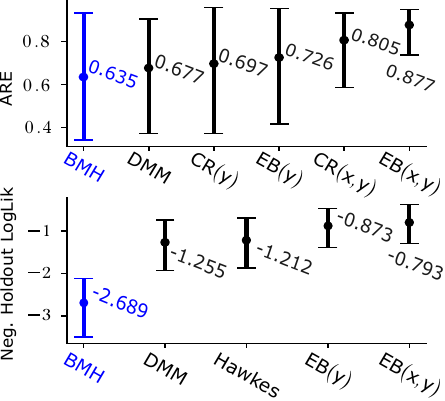}
        \caption{}
    \end{subfigure}

    \caption{Predictive performance for (a) CNIX and (b) RNIX. The dots indicate the median and the error bars give the $25^{th}/75^{th}$ quantiles. We compare the \bmh with the DMM \cite{Kong2020}, EB \cite{Tan2021}, cascade-size (CR) models, and the joint HP.} \label{fig:predictive}
\end{figure}

In the lower half of \cref{tab:ablation}, we see that for $CNIX$ each additional model component improves the log-likelihood, and that the full model $\alpha(\vec{y}^{a}) + z(\vec{x}^{ac},\vec{y}^{a})$ has the best performance. 
For $RNIX$ we observe that the variant without the seed user follower count, i.e., $\theta(\vec{y}^{a})+z(\vec{y}^{a})$, has the best performance.
This finding suggests that in cascades related to reputable media articles, the seed user is not as influential in determining how long a cascade unfolds. In contrast, for controversial media articles, the seed user plays a significant role.
We posit this is because the more fringe messaging in controversial media spreads through topical social groups (like conspiracy theorists, QAnon sympathizers and far-right supporters) \cite{Booth2024,Johns2024}.
As a result, the first endorser is particularly important to legitimize content within the group.
This is in contrast with the publicizing of traditional media articles on social media, where the most important factor is the publisher's reputation.
In the bottom row of \cref{fig:predictive}(a) and \cref{fig:predictive}(b), we see that similar to the popularity prediction task, the \bmhk model outperforms all benchmarks on median performance for both datasets.

\section{What-If? Headline Style Profiling}
\label{sec5:whatif}
This section performs a counter-factual analysis to show that \bmh successfully captures the relationship between headline writing style (i.e. neutral, clickbait or inflammatory) and content popularity and half-life.
We run a `What-If?' experiment, taking headlines of different writing styles and using the trained \bmh models to infer how these headlines would perform under different publishers. 

We utilize $HEADLINES$, a dataset of $1,227$ article headlines collected using the news aggregation platform The Daily Edit \cite{Lee2024}. 
The headlines come from four topics (Top Stories, Australia, Finance, and Climate Change) and six media sources (Daily Telegraph, Sky News, Sunday Morning Herald, The Guardian, news.com.au). 
Each headline was examined and sorted into one of three categories based on its informational and emotional content: neutral (N=727), clickbait (N=438) and inflammatory (N=62).
Neutral headlines are detailed and appropriate, avoiding unnecessary information or emotive language, e.g. \textit{`Australia's top military officer in the UK speaks ahead of Queen's funeral.'} 
Clickbait lacks informational and/or emotive quality without being misleading or inflammatory, often designed to attract attention, e.g. \textit{`Bizarre sight spotted amid Aussie floods.'} Inflammatory headlines contain unnecessary details, often on serious topics, and may include inappropriate emotional language or details that reinforce negative stereotypes, e.g. \textit{`Absolutely disgraceful': AFL fans blasted.'}

We use the trained publisher-level \bmh models in \cref{sec:predictive} to predict performance of article headlines for each publisher in $CNIX$ and $RNIX$: expected cascade size (\cref{eq:coldstartbayes}) (setting $\hat{C}_\rho=1$) and half-life (\cref{eq:halflifebayes}). We use the variants that include only item features (i.e. $\alpha(\vec{y}^a)+z(\vec{y}^a)$ for \bmhp and $\theta(\vec{y}^a)+z(\vec{y}^a)$ for \bmhk ) since cascade features are not available in this counter-factual setting.

\subsubsection{Results.}

\begin{figure}[t!]
    \centering
    \begin{subfigure}[t]{0.3\textwidth}
        \centering
        \includegraphics[width=\textwidth]{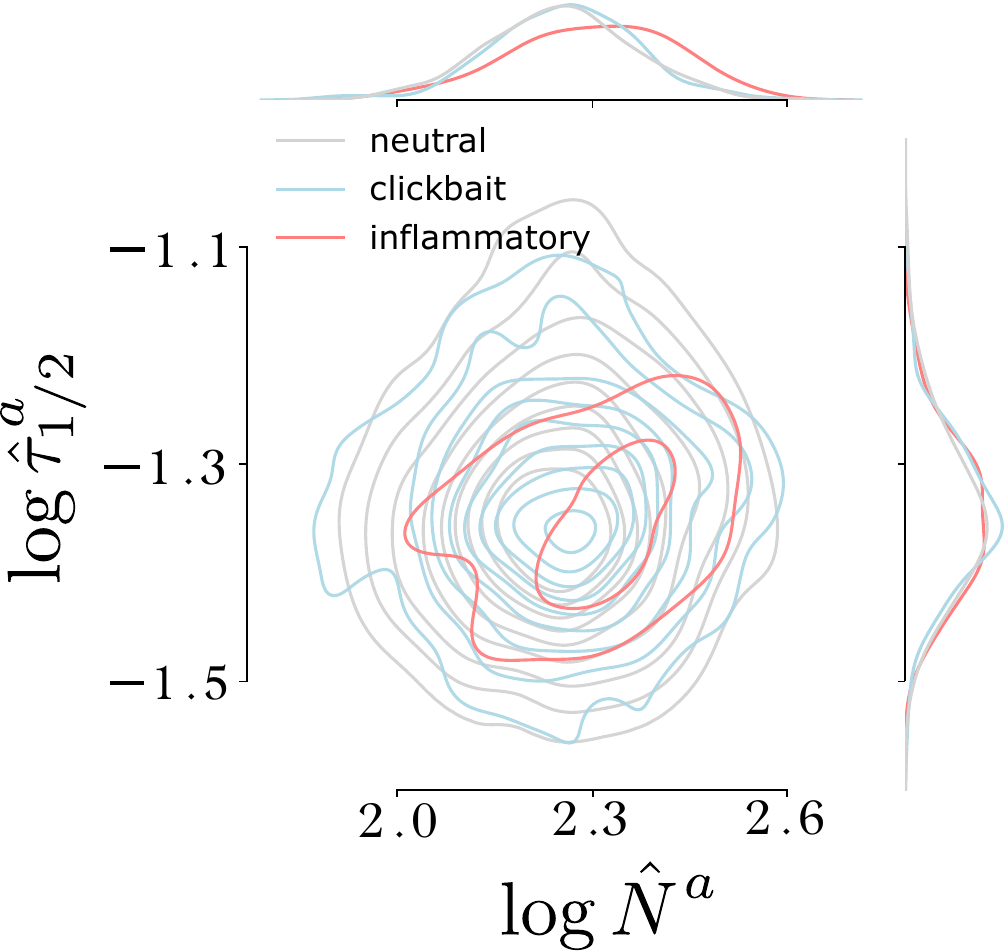}
        \caption{}
        \label{subfig:newscomau-contour}
    \end{subfigure}
    \hfill
    \begin{subfigure}[t]{0.33\textwidth}
        \centering
        \includegraphics[width=\textwidth]{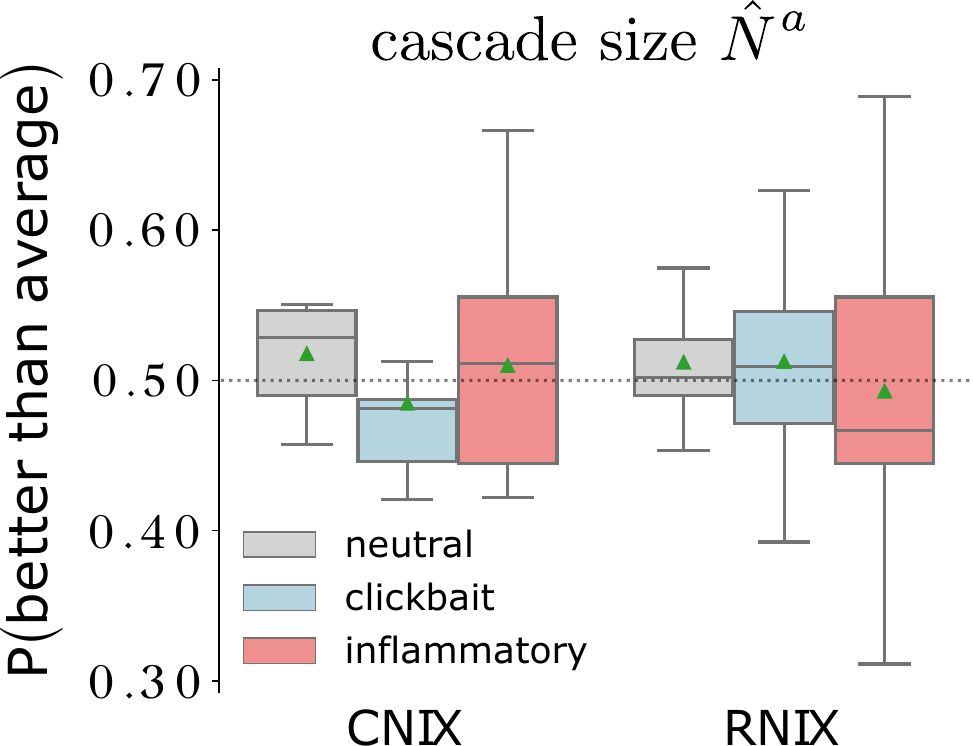}
        \caption{}
        \label{subfig:popularity-boxplots}
    \end{subfigure}
    \hfill
    \begin{subfigure}[t]{0.32\textwidth}
        \centering
        \includegraphics[width=\textwidth]{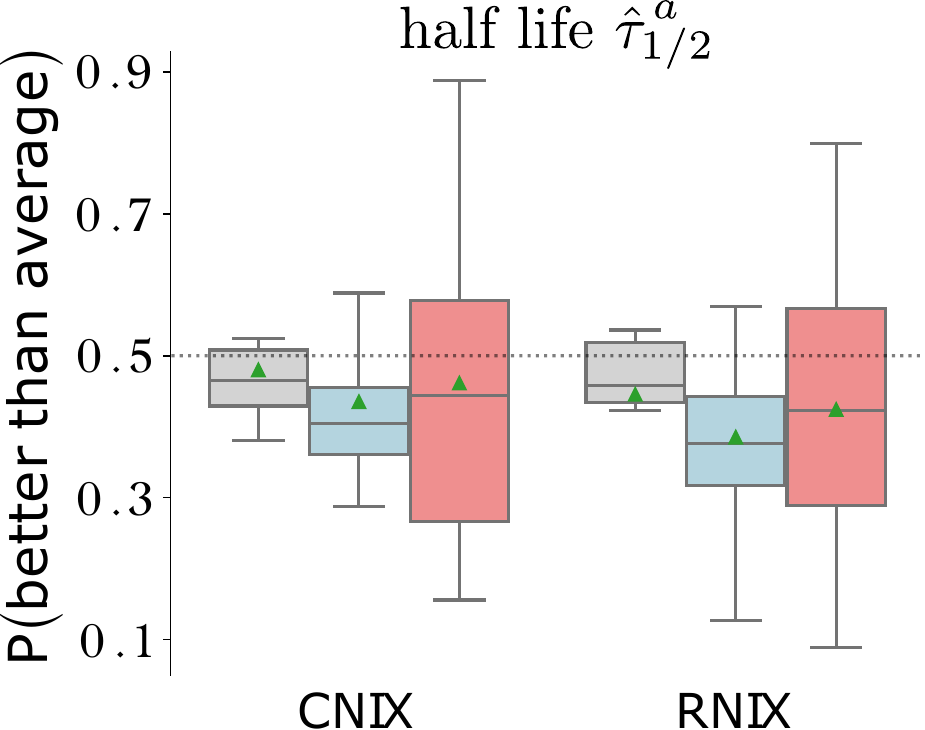}
        \caption{}
        \label{subfig:halflife-boxplots}
    \end{subfigure}
    \caption{(a) Distribution of predicted half-life $\log \hat{\tau}_{1/2}^a$ vs. cascade size $\log \hat{N}^{a}$ for each article in $HEADLINES$ using the \url{news.com.au} \bmh model. (b and c) Probability that an article performs better than the publisher average, for each headline style across $CNIX$ and $RNIX$: (b) cascade size $\hat{N}^{a}$; (c) half life $\hat{\tau}_{1/2}^a$.} 
    \label{fig:sec5mockup}
\end{figure}

We apply the trained the \bmhp/-K models of each publisher $\rho$ in $\{CNIX, RNIX\}$ to each of the $1,227$ article headlines in $HEADLINES$ to infer the article's peformance if it were published under $\rho$. 
We summarize the predictions with a publisher-level performance heatmap ($\log \hat{N}^{a}$ vs. $\log \hat{\tau}_{1/2}^a$), where we differentiate the performance of neutral, clickbait and inflammatory headlines by aggregating the predictions of each headline style as contour plots.
\cref{subfig:newscomau-contour} exemplifies the performance heatmap for the $RNIX$ publisher \url{news.com.au}. 
For this news source, we see that inflammatory headlines appear to have much higher popularity than neutral or clickbait headlines, while there is not much difference in half-life across headline styles.
This is somewhat expected, as this publisher is known for its tabloid tendencies, focusing on ``celebrity gossip, travel, lifestyle, sport, business, technology, money, and real estate'', according to Media Bias Fact Check (MBFC) \cite{Check2024}.
MBFC also rates its factual reporting as ``MOSTLY FACTUAL'' due to the occasional use of poor sources.
We observe differences in the patterns for the headline styles across publishers (see Section 4.1 of the Online Appendix \cite{appendix}), implying that effective headlines for one publisher might not be effective for another, and that the \bmh model learns these differences. 

To summarise the differences across the categories $CNIX$ and $RNIX$, we compute the probability that each headline performs better -- has a larger predicted cascade size or longer predicted half-life based on the \bmh -- than the publisher average based on the publisher's historical data. 
In \cref{subfig:popularity-boxplots,subfig:halflife-boxplots} we show the distribution of these probabilities for each category and headline style.

We have three observations for the popularity probabilities in \cref{subfig:popularity-boxplots}. First, we see that for $CNIX$, neutral headlines are effective (i.e. median better-than-average probability $>50\%$). 
In contrast, clickbait headlines are ineffective (i.e. median better-than-average probability $<50\%$). 
We link this result to the known inverse U-shaped relationship between clickbait volume and audience engagement \cite{Zhang2020}, where too little or too much clickbait leads to suboptimal attention, suggesting the existence of a \textit{sweet spot} for clickbait use.
The over-prevalence of clickbait in controversial media outlets results in clickbait fatigue among readers \cite{Lischka2023}, leading to diminished effectiveness of clickbait headlines observed in \cref{subfig:popularity-boxplots}.

Second and interestingly, 
we see that for $RNIX$ clickbait tends to perform better than neutral headlines.
This is explained by Rony et al~\cite{rony2017}, who show that traditional news-oriented media consist of only 22\% clickbait headlines while unreliable media consists of 39\% clickbait based on a large sample of headlines.
Since reputable media publishers have lower clickbait volume than controversial outlets, they are closer to the sweet spot for clickbait usage, retaining its effectiveness for drawing audience engagement. 
We do see a larger variance for clickbait for $RNIX$ compared to $CNIX$, suggesting that clickbait effectiveness is inconsistent and may not resonate universally, linking to the fact that clickbait strategies are only successful with certain audience segments \cite{Mukherjee2022}.

Third, we observe large variance of performance for inflammatory headlines in both categories, indicative of the polarizing nature of this headline style. 
Inflammatory headlines tend to perform better in controversial outlets.

For the half-life probabilities (\cref{subfig:halflife-boxplots}), we see similar results, except that neutral headlines in both categories have higher half-life than clickbait, demonstrating the ephemerality of clickbait \cite{Liao2020} irrespective of where it is published.

\section{Conclusion and Future Work}

This paper proposes the Bayesian Mixture Hawkes (\bmh) model, a hierarchical mixture model of Hawkes processes capable of learning the influence of item- and cascade-level features on spread dynamics. We demonstrate the applicability of the \bmh model on two retweet cascade datasets that reference articles from reputable and controversial online news sources and show that the \bmh model outperforms benchmark models in cold-start popularity prediction and temporal profile generalization performance. We apply the trained \bmh models to a dataset of article headlines written in different headline styles and show differences in performance of headline styles across reputable and controversial outlets. 

\pioc{\subsubsection{Limitations and Future Work.} 
We use the Hawkes process as the building block of the \bmh model since it does not require the branching structure of diffusion cascades for inference. This choice is driven by data limitations on Twitter, where the branching structure of content shares is not accessible.

We propose two improvements. 
First, the \bmh model assumes that $\alpha$ and $\boldsymbol{\Theta}$ depend only on cascade- and content-level features. 
We can allow $\alpha$ and $\boldsymbol{\Theta}$ to vary per event by including event-level features, which can be achieved by using the parametric Hawkes process \cite{Li2014} or Tweedie-Hawkes \cite{Li2020}. 
Second, the \bmh model assumes a fixed number of popularity/kernel classes, obtained empirically by pre-fitting with the DMM. 
We can learn the manifest number of components directly from the data by assuming an infinite number of components via nonparametric Bayesian methods, such as using a Dirichlet Process prior \cite{Navarro2006}. 

We aim to develop the \bmh model as a cold-start headline optimization tool by combining it with generative AI (e.g. ChatGPT \cite{openai2023}). The system would work in a `generate-then-evaluate' loop, where headlines are generated automatically by ChatGPT and then we apply the \bmh model to rank the generations.}

\begin{credits}
    \subsubsection{\ackname} This work was partially funded by the Australian Department of Home Affairs, the Defence Science and Technology Group, the Defence Innovation Network and the Australian Academy of Science.

    \subsubsection{\discintname}
    The authors have no competing interests to declare that are relevant to the content of this article.
\end{credits}
    
\bibliographystyle{splncs04}
\bibliography{mybibliography}

\begin{thebibliography}{10}
\providecommand{\url}[1]{\texttt{#1}}
\providecommand{\urlprefix}{URL }
\providecommand{\doi}[1]{https://doi.org/#1}

\bibitem{Bakshy2011}
Bakshy, E., Hofman, J.M., Mason, W.A., Watts, D.J.: Everyone's an influencer:
  quantifying influence on twitter. In: WSDM 2011

\bibitem{Bao2016}
Bao, P.: Modeling and predicting popularity dynamics via an influence-based
  self-excited hawkes process. In: CIKM 2016

\bibitem{Booth2024}
Booth, E., Lee, J., Rizoiu, M.A., Farid, H.: {Conspiracy, misinformation,
  radicalisation: understanding the online pathway to indoctrination and
  opportunities for intervention}. Journal of Sociology  (feb 2024)

\bibitem{Borel1942}
Borel, E.: Sur l'emploi du theoreme de Bernoulli pour faciliter le calcul d'une
  infinite de coefficients. CR Acad. Sci. Paris (1942)

\bibitem{appendix}
Calderon, P., Rizoiu, M.A.: Appendix: What drives online popularity: Author,
  content or sharers? \url{https://arxiv.org/pdf/2406.03390.pdf#page=19} (2024)

\bibitem{Carpenter2017}
Carpenter, B., Gelman, A., Hoffman, M.D., Lee, D., Goodrich, B., Betancourt,
  M., Brubaker, M., Guo, J., Li, P., Riddell, A.: Stan: A probabilistic
  programming language. Journal of statistical software  \textbf{76}(1) (2017)

\bibitem{Pewresearch}
Center, P.R.: Pew research center (2023),
  \url{https://www.pewresearch.org/journalism/fact-sheet/social-media-and-news-fact-sheet/}

\bibitem{Ghassemi2022}
Ghassemi, M., Dalmasso, N., Lamba, S., Potluru, V., Balch, T., Shah, S.,
  Veloso, M.: Online learning for mixture of multivariate hawkes processes. In:
  ICAIF 2022

\bibitem{Gomez2011}
Gomez-Rodriguez, M., Balduzzi, D., Sch\"{o}lkopf, B.: Uncovering the temporal
  dynamics of diffusion networks. In: ICML 2011 (2011)

\bibitem{Hawkes1971}
Hawkes, A.G.: Spectra of some self-exciting and mutually exciting point
  processes. Biometrika  \textbf{58}(1),  83--90 (1971)

\bibitem{Johns2024}
Johns, A., Bailo, F., Booth, E., Rizoiu, M.A.: Labelling, shadow bans and
  community resistance: did meta's strategy to suppress rather than remove
  covid misinfo and conspiracy theory on facebook slow the spread? Media
  International Australia

\bibitem{Kong2020}
Kong, Q., Rizoiu, M.A., Xie, L.: Describing and predicting online items with
  reshare cascades via dual mixture self-exciting processes. In: CIKM 2020
  (2020)

\bibitem{Lee2024}
Lee, J., Booth, E., Farid, H., Rizoiu, M.A.: {Misinformation is not about Bad
  Facts: An Analysis of the Production and Consumption of Fringe Content} (mar
  2024), \url{http://arxiv.org/abs/2403.08391}

\bibitem{Li2014}
Li, L., Zha, H.: Learning parametric models for social infectivity in
  multi-dimensional hawkes processes. In: AAAI 2014

\bibitem{Li2020}
Li, T., Ke, Y.: Tweedie-hawkes processes: Interpreting the phenomena of
  outbreaks. In: Proceedings of the AAAI Conference on Artificial Intelligence
  (2020)

\bibitem{Liao2020}
Liao, Y., Wang, S., Han, E., Lee, J., Lee, D.: Characterization and early
  detection of evergreen news articles. In: ML and Knowledge Discovery in
  Databases (2020)

\bibitem{Lischka2023}
Lischka, J., Garz, M.: Clickbait news \& algorithmic curation: Game theory
  framework of the relation bet. journalism, users and platforms. New Media \&
  Society

\bibitem{Ma2016}
Ma, J., Gao, W., Mitra, P., Kwon, S., Jansen, B.J., Wong, K.F., Cha, M.:
  Detecting rumors from microblogs with recurrent neural networks. In: IJCAI
  2016 (2016)

\bibitem{Check2024}
{Media Bias Fact Check}: {News.com.au -- Bias and Credibility} (2024),
  \url{https://mediabiasfactcheck.com/news-com-au/}

\bibitem{Mishra2016}
Mishra, S., Rizoiu, M.A., Xie, L.: Feature driven and point process approaches
  for popularity prediction. In: CIKM 2016. pp. 1069--1078

\bibitem{Mukherjee2022}
Mukherjee, P., Dutta, S., De~Bruyn, A.: Did clickbait crack the code on
  virality? Journal of the Academy of Marketing Science  \textbf{50}(3),
  482--502 (2022)

\bibitem{Navarro2006}
Navarro, D.J., Griffiths, T.L., Steyvers, M., Lee, M.D.: Modeling individual
  differences using dirichlet processes. Journal of mathematical Psychology
  \textbf{50}(2),  101--122 (2006)

\bibitem{openai2023}
OpenAI: Chatgpt (2023), \url{https://openai.com/chatgpt}, software tool

\bibitem{Parikh2019}
Parikh, S.B., Patil, V., Makawana, R., Atrey, P.K.: Towards impact scoring of
  fake news. In: MIPR 2019. IEEE

\bibitem{Pedregosa2011}
Pedregosa, F., et.al.: Scikit-learn: Machine learning in {P}ython. Journal of
  Machine Learning Research  \textbf{12},  2825--2830 (2011)

\bibitem{Remiers2019}
Reimers, N., Gurevych, I.: Sentence-bert: Sentence embeddings using siamese
  bert-networks. In: EMNLP 2019

\bibitem{Rizoiu2017}
Rizoiu, M.A., Xie, L., Sanner, S., Cebrian, M., Yu, H., {Van Hentenryck}, P.:
  {Expecting to be HIP: Hawkes Intensity Processes for Social Media}. In: WWW
  2017

\bibitem{rony2017}
Rony, M.M.U., Hassan, N., Yousuf, M.: Diving deep into clickbaits: Who use them
  to what extents in which topics with what effects? (2017)

\bibitem{Sharma2019}
Sharma, A., Ghosh, A., Fiterau, M.: Generative sequential stochastic model for
  marked point processes. In: ICML Time Series Workshop (2019)

\bibitem{Tan2021}
Tan, W.H., Chen, F.: Predicting the popularity of tweets using internal and
  external knowledge: an empirical bayes type approach. AStA  \textbf{105}(2),
  335--352 (2021)

\bibitem{cmdstanpy}
Team, S.: Cmdstanpy (0.9.76). \url{https://pypi.org/project/cmdstanpy} (2023)

\bibitem{Tsagkias2009}
Tsagkias, M., Weerkamp, W., de~Rijke, M.: Predicting the volume of comments on
  online news stories. In: CIKM 2009 (2009)

\bibitem{Zhang2020}
W, Z., W, D., Y, B., et~al.: Seeing is not always believing: an exploratory
  study of clickbait in wechat. Internet Research  \textbf{30}(3),  1043--1058
  (2020)

\bibitem{Xu2017}
Xu, H., Zha, H.: A dirichlet mixture model of hawkes processes for event
  sequence clustering. Advances in neural information processing systems
  \textbf{30} (2017)

\bibitem{Zhao2015}
Zhao, Q., Erdogdu, M.A., He, H.Y., Rajaraman, A., Leskovec, J.: Seismic: A
  self-exciting point process model for predicting tweet popularity. In: SIGKDD
  2015

\end{thebibliography}



\section*{Supplementary material}

Table 1: Table of Notation.

| Parameter | Interpretation | Real-World Mapping |
|---|---|---|
| *Source-Level* | | |
| $\rho$ | source of items | news publisher |
| $\mathcal{A}$ | set of items produced by $\rho$ | news articles from publisher $\rho$ |
| $f_\rho(\cdot)$ | follower count distribution | |
| $\hat{C}_\rho$ | cascade count estimate | |
| *Item-Level* | | |
| $a \in \mathcal{A}$ | item produced by $\rho$ | news article |
| $\mathbb{H}^a$ | set of cascades related to item $a$ | retweet cascades for article $a$ |
| $\vec{y}^a$ | item-level features | headline embedding for article $a$ |
| $N^a$ | item popularity | overall tweet count for article $a$ |
| $\tau^a_{1/2}$ | content half-life | |
| *Cascade-Level* | | |
| $\mathcal{H}^{ac} \in \mathbb{H}^a$ | cascade related to item $a$ | retweet cascade for article $a$ |
| $\vec{x}^{ac}$ | cascade-level features | follower count of seed user |
| $N^{ac}$ | cascade size | |
| $\tau^{ac}_{1/2}$ | cascade half-life | |
| $\mathcal{T}^{ac}$ | interevent-time distribution | |
| $\alpha^{ac}$ | Hawkes branching factor | |
| $\boldsymbol{\Theta}^{ac}$ | Hawkes kernel parameters | |
| *BMH-P* | | |
| $K_\alpha$ | # of BMH-P mixture classes | |
| $\vec{\gamma}_{\alpha,k}$ | effect of $\vec{y}^a$ on center of class $k$ | |
| $\vec{\gamma}_{z_{\alpha,k}}$ | effect of $\vec{y}^a$ on membership probability of class $k$ | |
| $\delta^a_{\alpha,k}/\delta_{\alpha,k}$ | item-/ publisher-level baseline value of logit($\alpha$) for class $k$ | |
| $\vec{\beta}^a_{\alpha,k}/\vec{\beta}_{\alpha,k}$ | effect of $\vec{x}^{ac}$ on center of class $k$ | |
| $z^{ac}_{\alpha,k}$ | mem. probability for class $k$ | |
| $\delta_{z_{\alpha,k}}/\delta_{z_{\alpha,k}}$ | item-/ publisher-level mem. prob. softmax baseline for class $k$ | |
| $\vec{\beta}^a_{z_{\alpha,k}}/\vec{\beta}_{z_{\alpha,k}}$ | effect of $\vec{x}^{ac}$ on membership probability for class $k$ | |
| $\vec{p}^a_\alpha/\vec{p}_\alpha$ | item-/ pub.-level parameter vector | |
| $\Sigma_\alpha/\Omega_\alpha$ | cov./ corr. matrix for $\vec{p}_\alpha$ | |
| *BMH-K* | | |
| $K_{\boldsymbol{\Theta}}$ | # of BMH-K mixture classes | |
| $\vec{\gamma}_{\theta,k}$ | effect of $\vec{y}^a$ on center of class $k$ | |
| $\vec{\gamma}_{z_{\boldsymbol{\Theta},k}}$ | effect of $\vec{y}^a$ on membership probability of class $k$ | |
| $\delta^a_{\theta,k}/\delta_{\theta,k}$ | item-/ publisher-level baseline value of log($\theta$) for class $k$ | |
| $\vec{\beta}^a_{\theta,k}/\vec{\beta}_{\theta,k}$ | effect of $\vec{x}^{ac}$ on center of class $k$ | |
| $z^{ac}_{\boldsymbol{\Theta},k}$ | mem. probability for class $k$ | |
| $\delta^a_{z_{\boldsymbol{\Theta},k}}/\delta_{z_{\boldsymbol{\Theta},k}}$ | item-/ publisher-level mem. prob. softmax baseline for class $k$ | |
| $\vec{\beta}^a_{z_{\boldsymbol{\Theta},k}}/\vec{\beta}_{z_{\boldsymbol{\Theta},k}}$ | effect of $\vec{x}^{ac}$ on membership probability for class $k$ | |
| $\vec{p}^a_{\boldsymbol{\Theta},k}/\vec{p}_{\boldsymbol{\Theta},k}$ | item-/ pub.-level kernel parameter baseline values for class $k$ | |
| $\vec{p}^a_{z_{\boldsymbol{\Theta}}}/\vec{p}_{z_{\boldsymbol{\Theta}}}$ | item-/ pub.-level membership probability parameters for class $k$ | |
| $\Sigma_{\boldsymbol{\Theta},k}/\Omega_{\boldsymbol{\Theta},k}$ | cov./ corr. matrix for $\vec{p}_{\boldsymbol{\Theta},k}$ | |
| $\Sigma_{z_{\boldsymbol{\Theta}}}/\Omega_{z_{\boldsymbol{\Theta}}}$ | cov./ corr. matrix for $\vec{p}_{z_{\boldsymbol{\Theta}}}$ | |

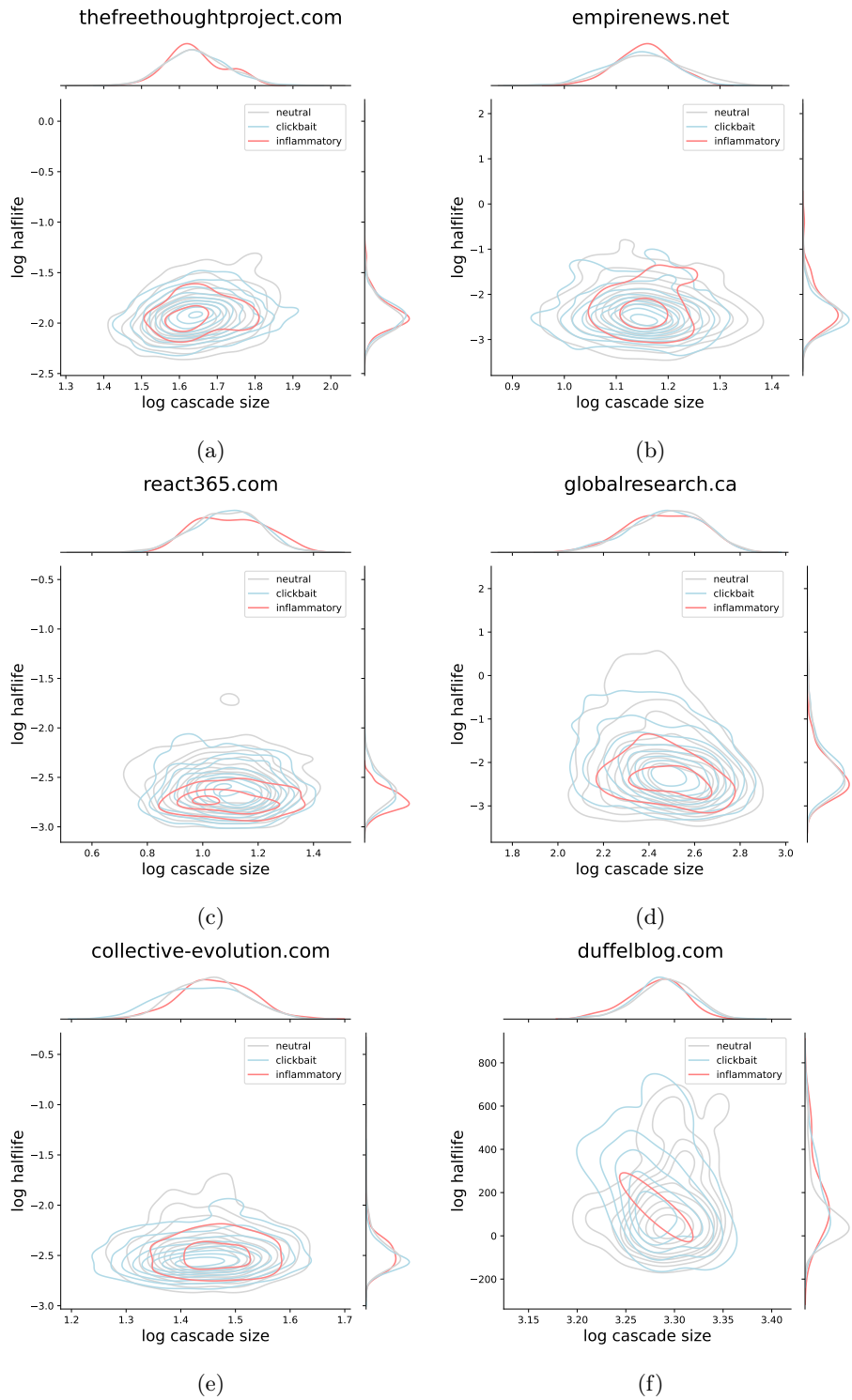

Fig. 3: Peformance heatmaps for a selection of CNIX publishers.

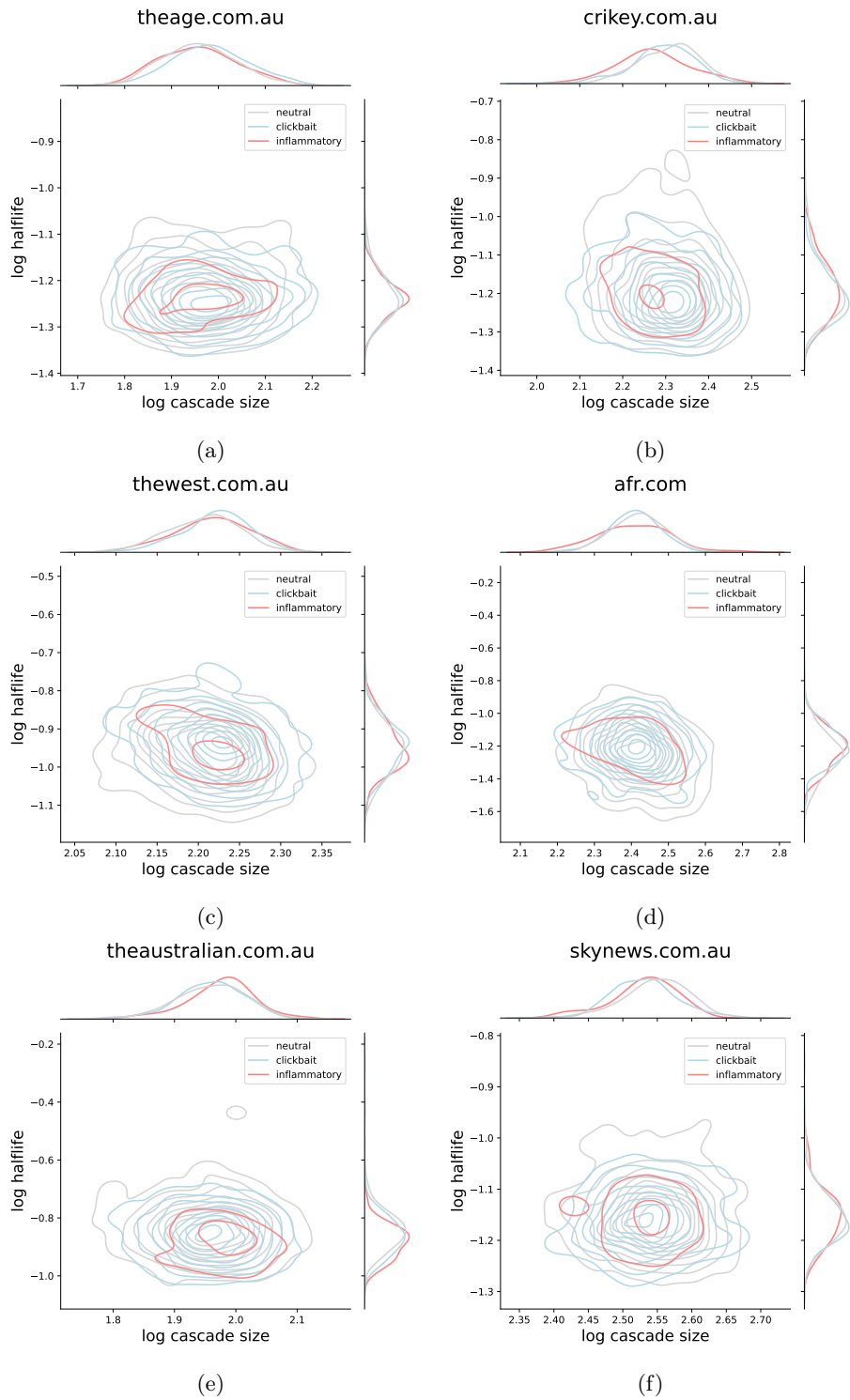

Fig. 4: Peformance heatmaps for a selection of RNIX publishers.



\end{document}